\documentclass[runningheads]{styles/llncs}

\usepackage{styles/eccv}

\usepackage{styles/eccvabbrv}

\usepackage{graphicx}
\usepackage{booktabs}

\usepackage[accsupp]{axessibility}  

\usepackage{hyperref}

\usepackage{orcidlink}

\usepackage{ifthen}

\newcommand{\termdeltapred}[1]{%
  \ifthenelse{\equal{#1}{norm}}%
    {prediction probability shifts}%
    {%
      \ifthenelse{\equal{#1}{cap}}%
        {Prediction probability shifts}%
        {Prediction Probability Shifts}%
    }%
}

\newcommand{\termproto}[1]{%
  \ifthenelse{\equal{#1}{norm}}%
    {abstract prototype}%
    {%
      \ifthenelse{\equal{#1}{cap}}%
        {Abstract prototype}%
        {Abstract Prototype}%
    }%
}

\begin{document}

\title{Why Does It Look There? Structured Explanations for Image Classification} 

\titlerunning{Why Does It Look There?}

\author{Jiarui Li\inst{1}\orcidlink{0009-0001-1055-4424} \and
Zixiang Yin\inst{1}\orcidlink{0009-0004-8725-1933} \and
Samuel J Landry\inst{2}\orcidlink{0000-0002-4082-0543} \and
Zhengming Ding\inst{1}\orcidlink{0000-0002-6994-5278} \and
Ramgopal R. Mettu\inst{1*}\orcidlink{0000-0001-9479-9156}}

\authorrunning{J.~Li et al.}

\institute{Department of Computer Science, Tulane University, New Orleans LA 70118, USA \and
Department of Biochemistry and Molecular Biology, Tulane University School of Medicine, New Orleans LA 70112, USA \\
*Corresponding Author\\
\email{\{jli78,zyin,landry,zding1,rmettu\}@tulane.edu}}
    \maketitle

    
\begin{abstract}
  Deep learning models achieve remarkable predictive performance, yet their black-box nature limits transparency and trustworthiness. Although numerous explainable artificial intelligence (XAI) methods have been proposed, they primarily provide saliency maps or concepts (\ie, unstructured interpretability). Existing approaches often rely on auxiliary models (\eg, GPT, CLIP) to describe model behavior, thereby compromising faithfulness to the original models. We propose Interpretability to Explainability (I2X), a framework that builds structured explanations directly from unstructured interpretability by quantifying progress at selected checkpoints during training using prototypes extracted from post-hoc XAI methods (\eg, GradCAM). I2X answers the question of ``why does it look there'' by providing a structured view of both intra- and inter-class decision making during training. Experiments on MNIST and CIFAR10 demonstrate effectiveness of I2X to reveal prototype-based inference process of various image classification models. Moreover, we demonstrate that I2X can be used to improve predictions across different model architectures and datasets: we can identify uncertain prototypes recognized by I2X and then use targeted perturbation of samples that allows fine-tuning to  ultimately improve accuracy. Thus, I2X not only faithfully explains model behavior but also provides a practical approach to guide optimization toward desired targets.
\end{abstract}
    \section{Introduction}
Deep learning models have been widely adopted across numerous domains (\eg, natural language processing~\cite{vaswani2017attention}, neuroscience~\cite{yin2025mind}, and immunology~\cite{li2025rational,li2025tcr}) due to their strong predictive performance, especially in image classification tasks~\cite{wang2021comparative}. However, the inherent black-box nature of these models remains a major concern~\cite{arrieta2020explainable}. To develop trustworthy and transparent deep learning systems, explainable artificial intelligence (XAI) has become not only a central component of modern deep learning research~\cite{dwivedi2023explainable}, but also a necessity in critical domains such as computational biology, medicine, politics, and economics~\cite{arrieta2020explainable,chen2024applying}. XAI has also attracted industrial companies (\eg, Anthropic, OpenAI) interesting to build more reliable and robust models~\cite{gao2025weight, anthropic2025biology}.

\begin{figure}[t]
    \centering
    \includegraphics[width=\linewidth]{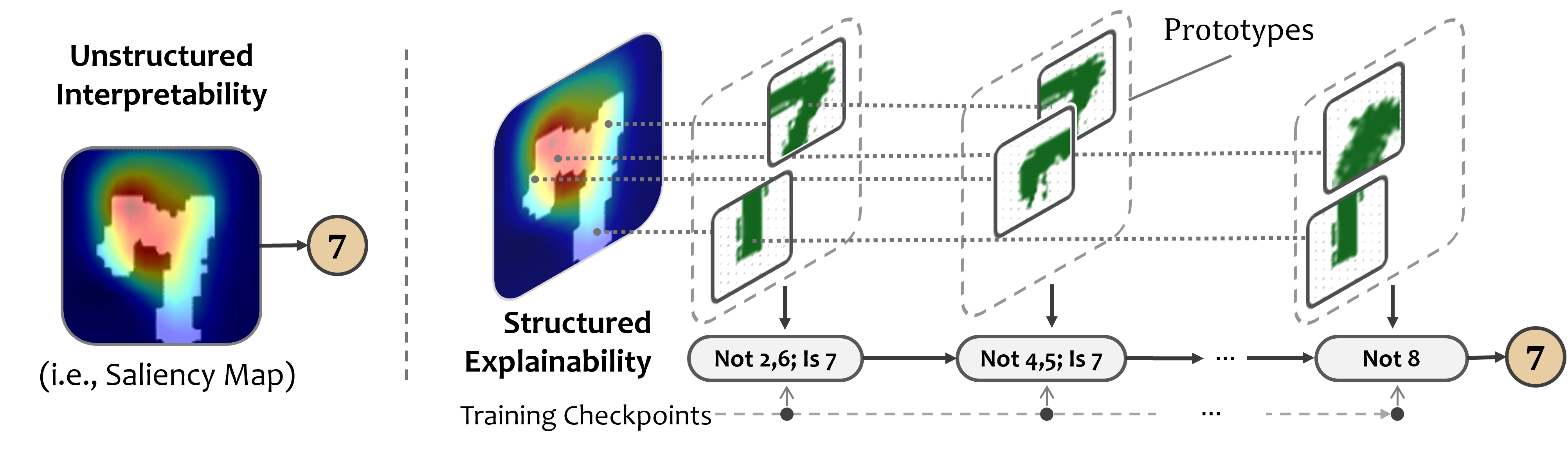}
    \caption{The difference between unstructured interpretability (\ie, with saliency maps) and our approach to structured explainability. I2X tracks model evolution across training checkpoints using a prototype-based representation.}
    \label{fig:ixdiff}
\end{figure}

A wide range of XAI methods have been proposed, and they can be categorized with respect to application scope (\eg, model-specific~\cite{wu2024token} vs.\ model-agnostic~\cite{chen2023algorithms}), explanation format (\eg, textual justifications~\cite{liu2025hybrid}, visualizations~\cite{achtibatattnlrp}), and whether they are post-hoc~\cite{selvaraju2017grad} or ante-hoc~\cite{koh2020concept,minh2022explainable}. Current XAI methods primarily generate visual interpretations (\eg, saliency maps~\cite{zhou2016learning}), collections of predicted or predefined human-interpretable concepts (\eg, concept vectors~\cite{kim2018interpretability}), or counterfactual examples~\cite{goyal2019counterfactual} to interpret model behavior. However, fundamental questions remain: \textbf{why} does a model focus on certain image regions, and \textbf{how} does the model organize these regions for learning and inference?
In other words, existing methods provide unstructured interpretability rather than structured explainability. Indeed, the concepts of interpretability and explainability have been discussed in philosophy of XAI~\cite{miller2019explanation}, and we adopt the view that explainability can be viewed as a post-hoc notion built upon interpretability~\cite{lipton2018mythos}. Ideally we would be able to provide a structured, causal representation for explanations (i.e. so that  ``an explanation is an assignment of causal responsibility''~\cite{halpern2005causes}). In our proposed framework we seek to describe the internal behavior of a model and thus focus on an assignment of responsibility of model behavior rather than causality.

\begin{figure}[t]
    \centering
    \includegraphics[width=\linewidth]{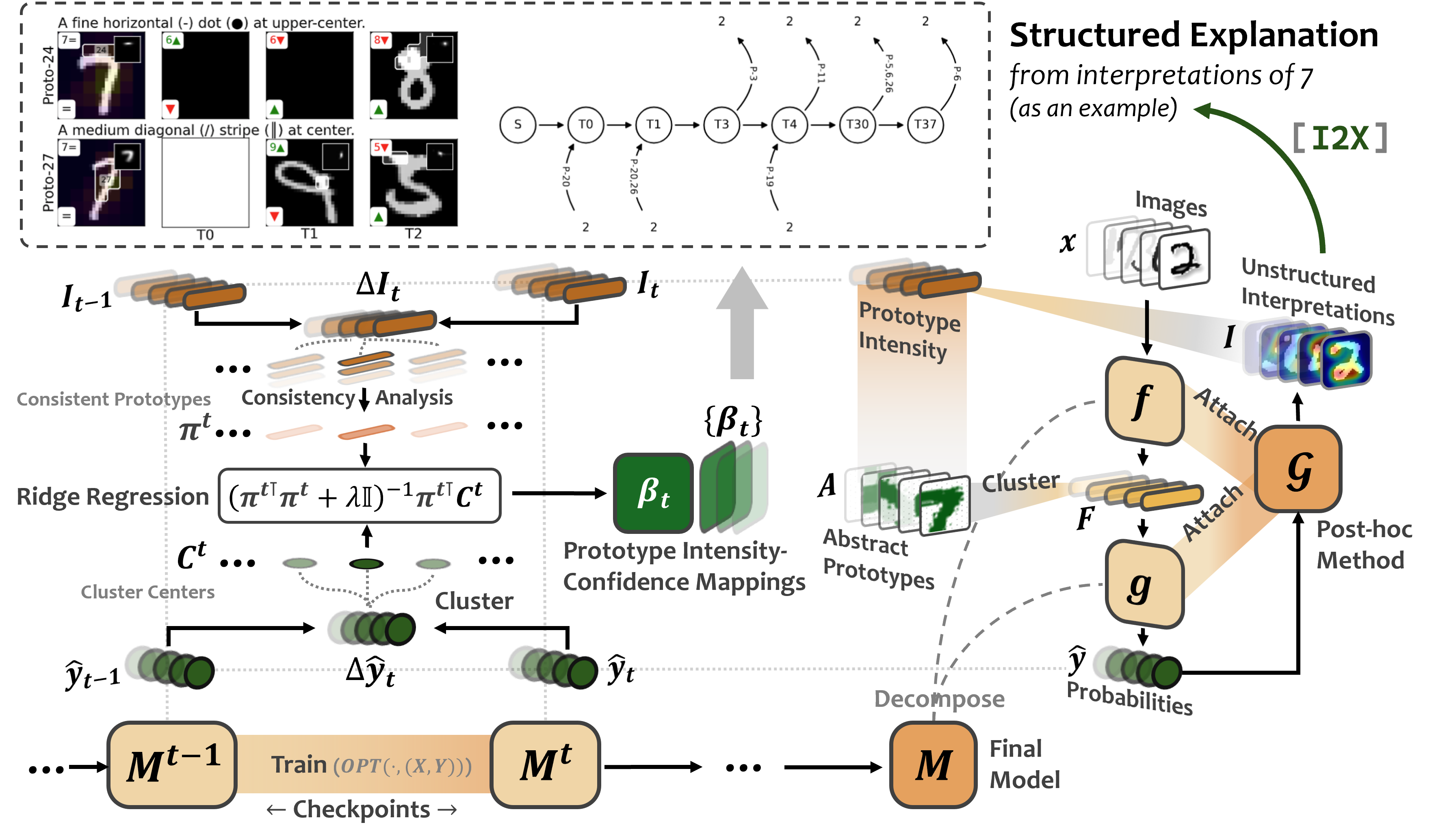}
    \caption{\textbf{Interpretability to Explainability (I2X).} I2X is a framework that builds structured explanation from evolutions of prototype intensity from saliency maps obtained by post-hoc method, and model confidences. I2X can systematically explain how a model infers and learns a predictive label and further utilize the resulting structured explanation to guide optimization toward the desired targets then improve performance.}
    \label{fig:pipeline}
\end{figure}

From this perspective, existing XAI approaches (\eg, concept learning~\cite{koh2020concept}, prototype learning~\cite{chen2019looks}, and saliency-based methods~\cite{selvaraju2017grad}) primarily provide \emph{unstructured interpretability} rather than \emph{structured explainability}. While some works attempt to introduce structured reasoning on top of interpretability (\eg, combining prototypes with reinforcement learning~\cite{kenny2023towards}) or utilize large language models to verbalize learned concepts (\eg, using HybridCBM ~\cite{liu2025hybrid}, GPT~\cite{achiam2023gpt}, CLIP~\cite{radford2021learning}). The resulting explanations are thus not intrinsic to the original model and can potentially create hallucinations from auxiliary models.
Methods such as DiffCAM~\cite{li2025diffcam} further organize saliency maps by contrasting activation patterns across samples or groups, offering improved faithfulness. However, they still do not yield structured explanations that explicitly characterize how the model organizes and utilizes these differences during inference or training. There has been work that follows the training influence~\cite{wang2025capturing} to rationalize how individual samples contribute to training, but it does not provide higher-level (\ie, feature) interpretation. Consequently, extracting the intrinsic structured explainability of a trained model remains an open and challenging problem.

We aim to enable the model itself to ``tell'' us its structured explanations; that is, how did it learn and perform inference within and across classes, based on a set of prototypes?  To this end, we propose Interpretability to Explainability (I2X), a framework that builds upon existing interpretability methods to construct \emph{structured explanations} by tracking model evolution across training checkpoints. Our approach consists of two main steps (as illustrated in~\cref{fig:ixdiff}): 
(1) construct a collection of model prototypes from unstructured interpretations (\ie, saliency maps) summarizing similar patterns, and 
(2) assign responsibility across the prototypes collection using model confidence across selected, incremental training checkpoints. Then, we can construct an explanation for a particular input by looking at the role of each prototype in the collection at each checkpoint in the final classification output.   

Our approach directly answers the question of \emph{``why does it look there''} by correlating model confidence and prototypes recognized from unstructured interpretation during training. We show that our method transforms unstructured post-hoc interpretations (\ie, saliency maps obtained by GradCAM~\cite{selvaraju2017grad}) into structured explanations, revealing the learning and inference trajectory of ResNet-50~\cite{he2016deep} trained on MNIST~\cite{deng2012mnist} and CIFAR10~\cite{krizhevsky2009learning}, and InceptionV3~\cite{szegedy2016rethinking} trained on MNIST. I2X illustrates how the model learns to identify a digit using prototypes by revealing how it is distinguished from other digits during training. This process enables perturbation of samples with uncertain prototypes that confuse the model, reducing confusing samples and improving performance. Furthermore, by understanding how the model infers a digit or object from specific abstract prototypes extracted from interpretations, we identify uncertain prototypes that hinder correct decisions and drive suboptimal optimization. Utilizing these insights, our framework enables targeted fine-tuning that steers optimization toward the desired target and improves performance, reducing the number of confused samples between the most confused classes by about 5 on MNIST and 23 on CIFAR-10, respectively.
    \section{Our Approach}

In this section, we introduce the I2X framework as shown in \cref{fig:pipeline}. First, we formalize the information extracted from the model confidence and interpretation. We then describe how this information is organized to derive prototypes and sample groups, enabling subsequent analysis. Next, we present how structured explanations are assembled by tracing the evolution of prototype intensity and model confidence during training at selected checkpoints.

\subsection{Preliminaries}
To simplify the formulation, we consider the image classification problem as pairs $(x, y)$, where 
$x\in \mathbb{R}^{H \times W \times 3}$ denotes an input image with a spatial resolution of $H\times W$. The one-hot vector
$y \in \{0,1\}^M$ represents the target label with $M$ classes. 
A classification model is defined as $\mathcal{M}: \mathbb{R}^{H\times W\times 3} \to \mathbb{R}^M$, where $\hat{y}=\mathcal{M}(x)\in[0,1]^M$ denotes the predicted class confidence. 

Following the standard structure of deep learning models, we decompose $\mathcal{M}$ into two components: a feature extractor $f: \mathbb{R}^{H\times W\times3} \to \mathbb{R}^{h\times w\times d}$ and a classifier head $g: \mathbb{R}^{h\times w\times d} \to \mathbb{R}^{M}$, such that $\mathcal{M} = g \circ f$. 
Here, $F=f(x)\in\mathbb{R}^{h\times w \times d}$ denotes the spatial latent feature representation, where $h$ and $w$ are the shape of the latent features, and $d$ is the number of dimensions.

An interpretation method $\mathcal{G}$ (\eg, CAM~\cite{zhou2016learning}) utilizes the model $\mathcal{M}$ and its prediction $\hat{y}$ to produce a saliency map for input $x$. 
This can be formulated as:
\begin{align}\label{eq:intp}
    I=\mathcal{G}(\mathcal{M}, \hat{y})\in [0,1]^{h \times w},
\end{align}
where $I \in [0,1]^{h \times w}$ shares the same shape of $F$ and can be upsampled to represent the saliency map over the input image.
To obtain saliency maps, various methods have been proposed for different model architectures, such as class activation maps (\eg, CAM~\cite{zhou2016learning}, GradCAM~\cite{selvaraju2017grad}, GradCAM++~\cite{chattopadhay2018grad}) for convolution neural networks (CNNs), and methods (\eg, TokenTM~\cite{wu2024token}, AttnLRP~\cite{achtibatattnlrp}, QCAI~\cite{li2025quantifying}) for transformers~\cite{vaswani2017attention}. 
As this paper focuses on CNNs, we adopt GradCAM to generate saliency maps for I2X.

We denote the training process as a mapping $\mathcal{M}_t=\text{OPT}\left(\mathcal{M}_{t-1}, \left(X,Y\right)\right)$, which optimizes a model $\mathcal{M}$ using paired data samples $(X,Y)$ to produce an optimized model from training iteration $t-1$ to iteration $t$ with initial untrained model  $\mathcal{M}_0$. The training can thus be written as a sequential optimization:
\begin{align}
    \mathcal{M}^t=\text{OPT}\left(\mathcal{M}^{t-1}, \left(X,Y\right)\right)=\cdots=\text{OPT}^{(t)}\left(\mathcal{M}^0, \left(X,Y\right)\right),
\end{align}
where OPT$(\cdot, (X,Y))$ is optimizing the given model checkpoint using data $(X,Y)$.
For the model $\mathcal{M}^t$, the prediction confidence is $\hat{y}^t = \mathcal{M}^t(X)$, and the corresponding interpretation in Eq.\eqref{eq:intp} is denoted by $I^t$. In addition, the final model (\ie, the model from the last training iteration $T$) is represented as $\mathcal{M}^T$ or $\mathcal{M}$.

\subsection{Abstract Prototypes}
We define \textit{abstract prototypes} as representative patterns summarizing groups of similar patterns across the learned data, distinguishing them from the ``prototypes'' in prototype learning~\cite{chen2019looks}. Throughout this paper, prototype refers to an abstract prototype.
To reduce the number of features learned by the model and simplify subsequent analysis, we first apply principal components analysis (PCA) and then $K$-Means to cluster all $N\times h\times w$ hidden feature vectors $\mathbf{F}=[F_1,F_2,\dots,F_N]\in\mathbb{R}^{(N\cdot h\cdot w)\times d}$ extracted by the feature extractor $f$ of the final model $\mathcal{M}$ using all $N$ training samples $x \in X$. This yields $K$ centroids as abstract prototypes and cluster assignments for all $h\times w$ feature vectors for the $i$-th sample, which can be represented as
\begin{align}
A_i=(a_1,a_2,\dots,a_{hw}),
\end{align}
where $a_j\in\{1,\dots,K\}$ denotes the cluster assignment corresponding to $i$-th feature vector, which also serve as indices of the abstract prototypes. After clustering, our desired prototypes correspond to the centroids and learned by $K$-Means, representing recurring patterns recognized by the model across the dataset. For a given image $x$, the corresponding cluster assignment $A\in\mathbb{R}^{h\cdot w}$ is aligned with the feature $F\in\mathbb{R}^{h\times w\times d}$ in position. Since $F$ represents the features of local patches of $x$, each patch can be described by its corresponding unit in $A$, linking image regions to prototypes.

\subsection{Prototype Intensity During Training}
For a training checkpoint $\mathcal{M}^t$ at step $t$, the unstructured interpretations (\ie, saliency maps) are obtained using the post-hoc method $\mathcal{G}$ in~\cref{eq:intp}, defined as $I^t = \mathcal{G}(\mathcal{M}^t, \hat{y}^t)\in\mathbb{R}^{h\times w},$
denoting the interpretation for a given input sample $x$.

For each input sample $x$, we summarize the associated interpretation into a prototype intensity representation $\mathbf{P}^t$ by aligning it to abstract prototypes as
\begin{align}\label{eq:proto_mig}
    \mathbf{P}^t=(P^t_1, P^t_2,\dots,P^t_K)\in\mathbb{R}^K \quad\text{with}\quad P^t_k=\frac{\sum_{j=1}^{hw}\mathbf{1}[a_j=k]\cdot\text{Flatten}(I^t_j)}{\sum_{j=1}^{hw}\mathbf{1}[a_j=k]}.
\end{align}
This formulation aggregates the interpretation values over spatial locations assigned to each prototype, effectively quantifying the activation strength (\ie, representation intensity) of each abstract prototype for the given sample. The change of prototype intensity $\Delta \mathbf{P}^t=\mathbf{P}^{t+1}-\mathbf{P}^{t}\in\mathbb{R}^{K}$ across two training checkpoints characterizes how the evidential prototypes selected by the model evolve as confidence change throughout training.

\subsection{Tracking Model Confidence With Prototypes}
As evidential prototype intensity evolves across training, the model confidence evolves accordingly. The confidence change between two consecutive checkpoints $t$ and $t+1$ is defined as $\Delta \hat{y}^t = \hat{y}^{t+1} - \hat{y}^t\in\mathbb{R}^{M}.$
To characterize common patterns of confidence change, we apply HDBSCAN~\cite{mcinnes2017hdbscan} to cluster the confidence change across two checkpoints for all samples and obtain their assignments $L^t$ with a total of $Q+1$ centroids:
\begin{align}\label{eq:pp_mig}
L^t = \text{HDBSCAN}(\Delta \hat{Y}^t)\in\{-1,1,2,\dots,Q\}^N,
\end{align}
where $\Delta\hat{Y}^t\in\mathbb{R}^{N\times M}$ is the collection of $\Delta \hat{y}^t$ for all $N$ training samples. HDBSCAN~\cite{campello2013density} is a density-based clustering method by constructing a hierarchy of clusters and selecting stable clusters based on density persistence, which can automatically determine the number of clusters, handle clusters with varying densities, and identify noise points (label as $-1$).
This operation groups samples exhibiting similar patterns of change in model confidence change patterns, which correspond to groups of samples sharing common changes in prototype intensity.

\subsection{Mapping Between Prototype Intensity and Model Confidence}
Based on the evolution of prototype intensity in~\cref{eq:proto_mig} and model confidence~\cref{eq:pp_mig}, we construct a mapping between them to quantify how prototype intensity change with confidence change across checkpoints.

Let $L^t = (L_1^t, \dots, L_{N}^t) \in \{-1,1,2,\dots,Q\}^N$ be a label sequence. Define the index groups corresponding to each label value as $S_q^t = \{i\in\{1,\dots,N\}|L^t=q\},$
for all $q \in \{-1,1,2,\dots,Q\}$. Each group $S_q^t$ has cardinality $|S_q^t|$ which specifies the number of samples assigned to cluster $q$.
For cluster $q$, the aggregated confidence change for cluster is computed as 
\begin{align}
    c_q^t = \frac{1}{|S_q^t|}\sum_{s\in S_q^t} \Delta \hat{y}^t_s \in \mathbb{R}^{M}.
\end{align}
Similarly, the corresponding prototype intensity evolution is defined as
\begin{align}
    \tilde{P}_q^t = \frac{1}{|S_q^t|}\sum_{s\in S_q^t} \text{ReLU}(\Delta P^t_s - \sigma) \in \mathbb{R}^{K},
\end{align}
where $\sigma \in [0,1]$ is a threshold that filters out less important prototypes conditioned on the given cluster of similar confidence change. This construction yields cluster-level prototype intensity and confidence consistently changing pair $C^t= (c_1^t,c_2^t,\dots,c_Q^t)$ and $\pi^t=(\tilde{P}_1^t,\tilde{P}_2^t,\dots,\tilde{P}_Q^t)$, enabling structured modeling of how prototype intensity changes drive changes in model confidence.

Secondly, we model the relationship between prototype intensity and model confidence changes using ridge regression:
\begin{align}
\beta^t = \left(\pi^{t\top}\pi^t + \lambda \mathbf{I}\right)^{-1} \pi^{t\top}C^t \in \mathbb{R}^{K\times M},
\end{align} 
where $\mathbf{I}$ is the identity matrix and $\lambda \ge 0$ is the ridge regularization parameter. The coefficient matrix $\beta^t$ quantifies how changes in prototype intensity drive the change of model confidence at training step $t$.

\subsection{Assembling Structured Explanations}
By aggregating $[\beta_t]_{t \in \{0, \dots,T\}}$ over the entire training process, we obtain a global view of how the model organizes evidential prototype intensity to support a specific class or to distinguish between classes by evolving correlated confidence. 
This aggregation transforms unstructured interpretations into structured explanations by revealing consistent prototype intensity and confidence change across training.
However, even when focusing on a single class, multiple prototypes may contribute to confidence change, making direct analysis complex. 
To simplify the analysis, we study each class from two complementary perspectives: 
(1) identifying shared prototypes that consistently support predictions across the class, and 
(2) detecting specialized prototypes that are primarily associated with a subset of samples and differentiate them from the overall class pattern.

To identify shared prototypes, we compare the prototype assignments of all samples within a class and select the prototypes that are present in every sample. 
To identify specialized prototypes, we first cluster the samples of a class based on the prototypes they represent. 
For each resulting cluster, we analyze the prototypes that appear within the cluster but are absent from the shared prototypes of the class. 
This distinction allows us to separate common evidential patterns from subgroup-specific features, facilitating more explainable and fine-grained analysis.
When analyzing prototype evolution for a specific class, we focus only on training iterations in which model confidence change significantly due to prototype intensity evolution. 
For these key prototypes, we further examine which class confidence exhibit inverse changes, as such opposing shifts are considered to respond to the corresponding prediction update at that step.
    \section{Results}
\begin{figure}[t]
    \centering
    \includegraphics[width=\linewidth]{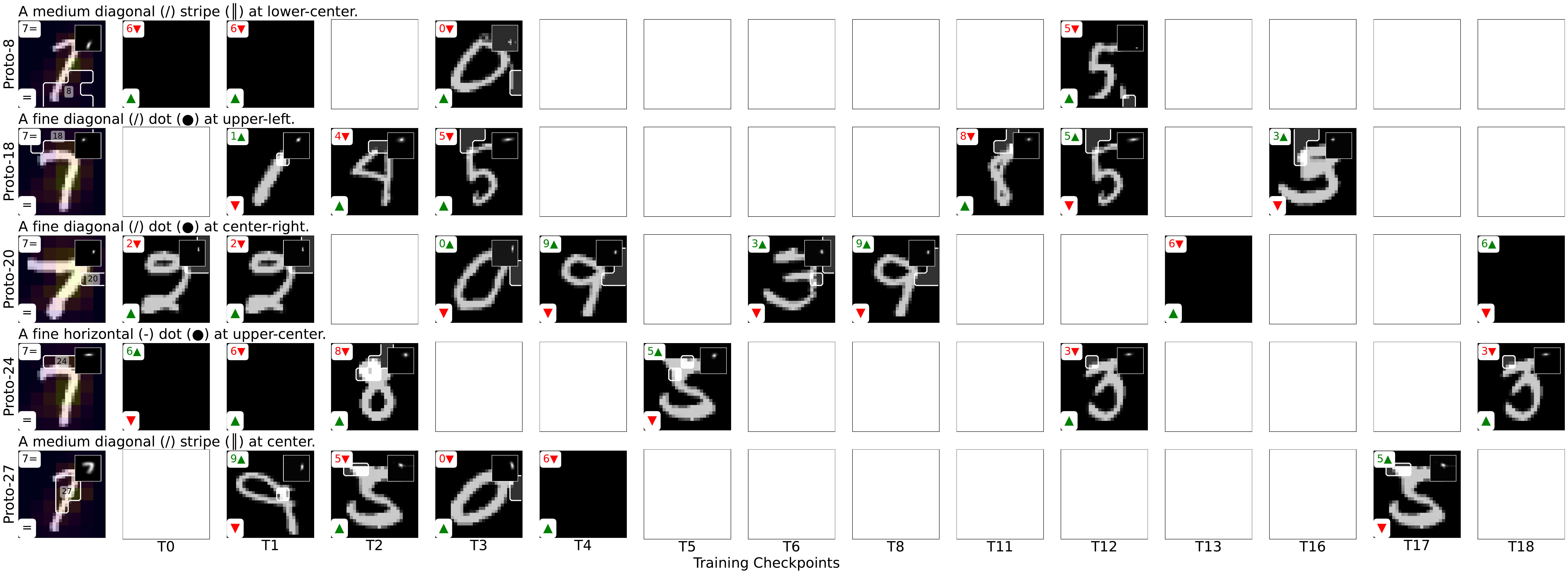}
    \includegraphics[width=\linewidth]{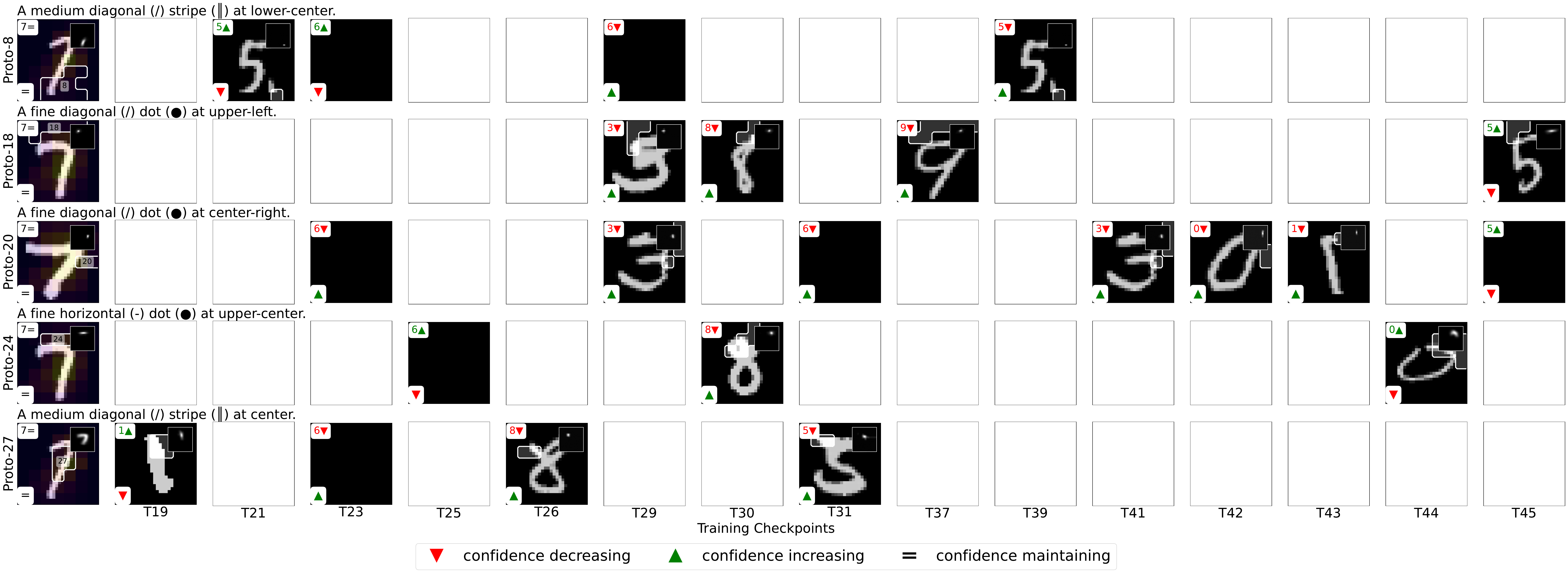}
    \caption{\textbf{Visualization of Prototype and Confidence Evolution of ResNet-50 on MNIST (shared prototypes of digit 7 as example).} 
    The first column shows the abstract prototypes of digit 7. Subsequent columns depict how sharing each prototype affects confidence for digit 7 (increase or decrease) and which competing class drives the change. White blanks denote no confidence change at that checkpoint, while black blanks indicate a change because 7 contains the prototype whereas another digit does not.}
    \label{fig:tlcmp:T2_7_shared}
\end{figure}
In this section, we show how I2X analyzes learning and inference mechanisms. We describe the experimental setup, present the structured explanation format, and use I2X to analyzing how ResNet-50 learns digit \emph{7} on MNIST. We further examine how randomness in data ordering affects learning strategies, using digits \emph{7} and \emph{2} to illustrate impacts on class interactions and performance. We then show how structured explanation guides fine-tuning by perturbing samples with uncertain prototypes to converge to their expected target. Finally, we extend the analysis to CIFAR-10 with ResNet-50 and to InceptionV3 on MNIST to validate generalizability across datasets and architectures.

\subsection{Experimental Setup}
Experiments are primarily conducted on ResNet-50~\cite{he2016deep}, with generalization validated on InceptionV3~\cite{szegedy2016rethinking}. 
Models are trained and analyzed on MNIST~\cite{deng2012mnist}, with additional evaluation on CIFAR10~\cite{krizhevsky2009learning}. 
Training on the full MNIST set uses a learning rate of $1\mathrm{e}{-3}$, 2 epochs, and batch size 128. 
The explanation dataset consists of $10\%$ randomly sampled training data (seed 42), and fine-tuning uses a learning rate of $1\mathrm{e}{-4}$. 
The number of K-Means clusters is set to 32 to balance diversity and simplicity. 
I2X analysis is performed every 40 iterations on MNIST and once per epoch on CIFAR10 due to slower convergence with 50 epochs.

\subsection{Structured Explanation for Digit Classification}

\begin{figure}[t]
    \centering
    \includegraphics[width=\linewidth]{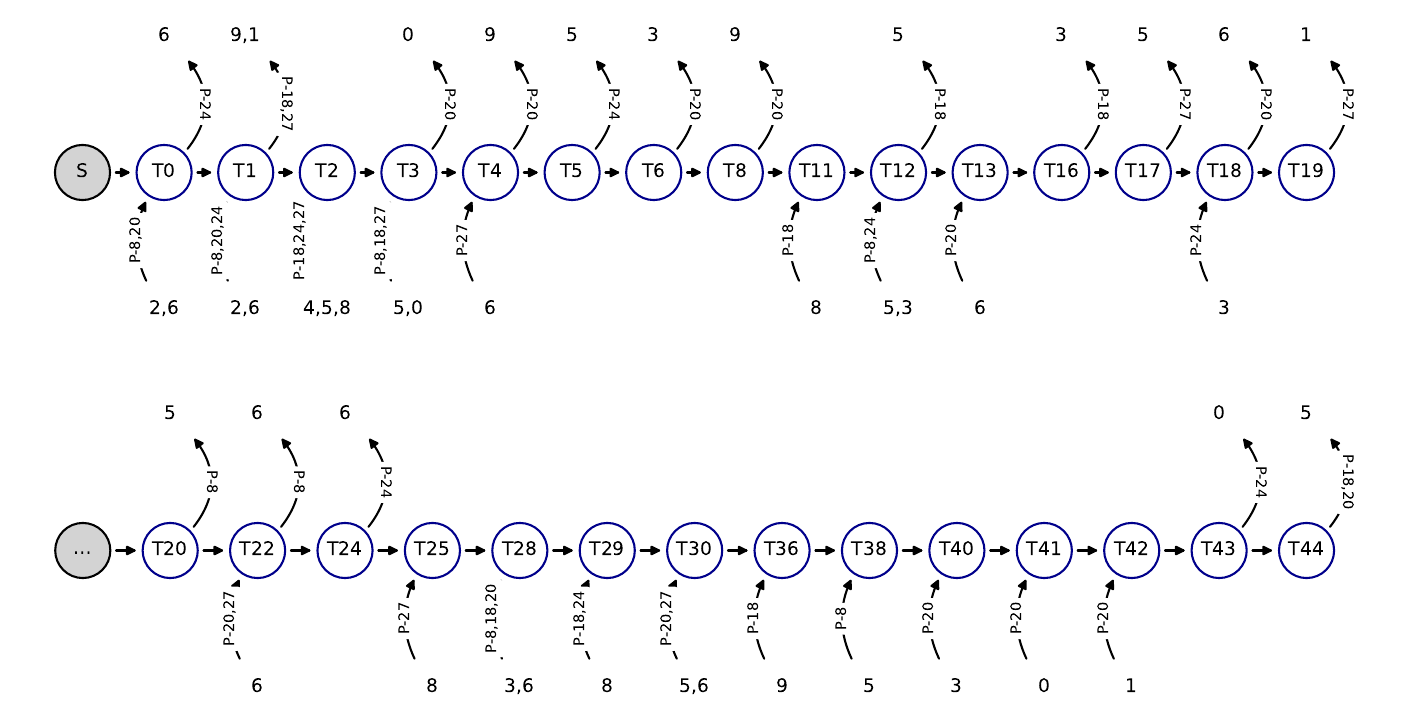}
    \caption{\textbf{Annotated Training Checkpoints for ResNet-50 on MNIST (shared prototypes of digit 7 as example).} This figure shows how the raw model evolves across training checkpoints. An edge $a \to b$ denotes that the model decreases confidence of class $a$ and increase confidence of class $b$. For example, $6,2 \to T0$ indicates that the model increases the prediction confidence of digit 7 (training checkpoint $T0$) while decreasing the confidence of digits 6 and 2. The labels on the arrows specify the prototypes responsible for each confidence change.}
    \label{fig:fsm:T2_7_shared}
\end{figure}

Using digit 7 as a case study, we demonstrate how the model learns to recognize 7 through two complementary visualization formats: 
(1) a visualization of prototype and confidence evolution, as shown in~\cref{fig:tlcmp:T2_7_shared}, and 
(2) an annotated training checkpoints figure, as shown in~\cref{fig:fsm:T2_7_shared}.

\cref{fig:tlcmp:T2_7_shared} and \cref{fig:fsm:T2_7_shared} show the model learning process based on the shared prototypes of class 7 that contribute to increasing confidence of digit 7. For clarity, we display only the training checkpoints at which the model confidence for class 7 changes. Five shared prototypes (\ie, prototypes 8, 10, 20, 24, and 27) are identified as contributing to these confidence changes, each corresponding to distinct structural components of digit 7.
We observe that the model first separates digits 2 and 6 from digit 7 using prototypes P-20 and P-8/P-24, respectively. 
Digit 6 lacks both the lower-center diagonal stroke (P-8) and the upper-center horizontal stroke (P-24), enabling the model to reliably distinguish 7 from 6 using these cues. 
For digit 2, only a small subset contains prototype P-20 (a center-right corner pattern). 
Since most instances of digit 7 exhibit this prototype while most 2 digits do not, the model uses P-20 as uncertain prototype evidence for 7.

\begin{figure}[t]
    \centering
    \includegraphics[width=\linewidth]{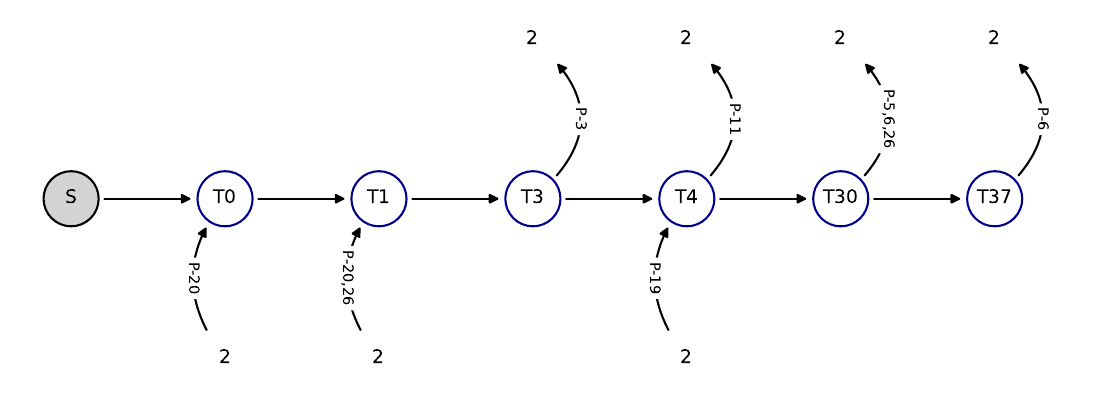}
    \caption{Annotated training checkpoints for ResNet-50 on MNIST illustrating all prototypes involved in distinguishing digit 7 from digit 2. Prototypes P-20 and P-19 are consistently assigned as discriminative evidence for digit 7. In contrast, prototype P-26 confuses model with its contribution alternating between digits 7 and 2 across training checkpoints, indicating uncertain evidence during training.}
    \label{fig:split:2_7:t2}
\end{figure}
The model separates digits 7 from other digits in a similar progressive manner, except for digits 1 and 9, which share multiple prototypes with digit 7. 
As a result, the model cannot identify a definitive distinguishing prototype between 7 and 1/9 until checkpoints T43 and T31, where it discovers P-20 and P-18 as uncertain prototype evidence, respectively.
This observation suggests that the model does not simultaneously differentiate a digit from all other classes. 
Instead, it first resolves classes with clear prototype discrepancies (\eg, digit 2, 6), and subsequently addresses more ambiguous cases (\eg, digit 1, 9) once sufficiently uncertain prototype evidence emerges.

\subsection{Distinguishing Between Two Digits}
To investigate how the model distinguishes between two digits, we take digits 7 and 2 as an illustrative example. 
As shown in~\cref{fig:split:2_7:t2}, prototype P-20 serves as reliable evidence for classifying a sample as digit 7, whereas P-26 behaves as a uncertain prototype that alternates its contribution between digits 7 and 2 during training.

When the model assigns P-19 as additional evidence for recognizing digit 7, it subsequently reassigns P-26 to support digit 2. 
This behavior indicates that the model does not rely on a single prototype for decision-making. 
Instead, predictions are formed through combinations of multiple prototypes. 
Such combinations include uncertain or shared prototypes, whose roles are dynamically determined based on the currently established prototype evidence.
This observation suggests that prototype utilization and composition are highly dependent on the order in which prototypes are selected and reinforced during training. 
Consequently, the randomness in training data ordering can significantly influence the model's optimization trajectory and ultimately affect its convergence and performance.

\begin{figure}[t]
    \centering
    \includegraphics[width=\linewidth]{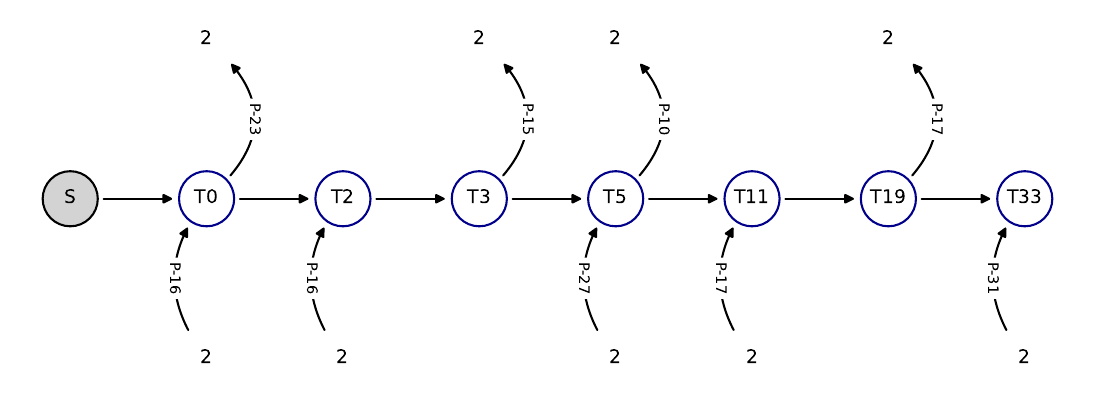}
    \caption{Annotated training checkpoints for ResNet-50 on MNIST illustrating how a re-trained model distinguishes digit 7 from digit 2. The figure demonstrates that, due to the randomness in training data order, different training runs result in distinct prototype-based inference strategies.}
    \label{fig:split:2_7:t1}
\end{figure}

\subsection{Analyzing Variations in Training Data Order}
\begin{figure}[t]
    \centering
    \includegraphics[width=\linewidth]{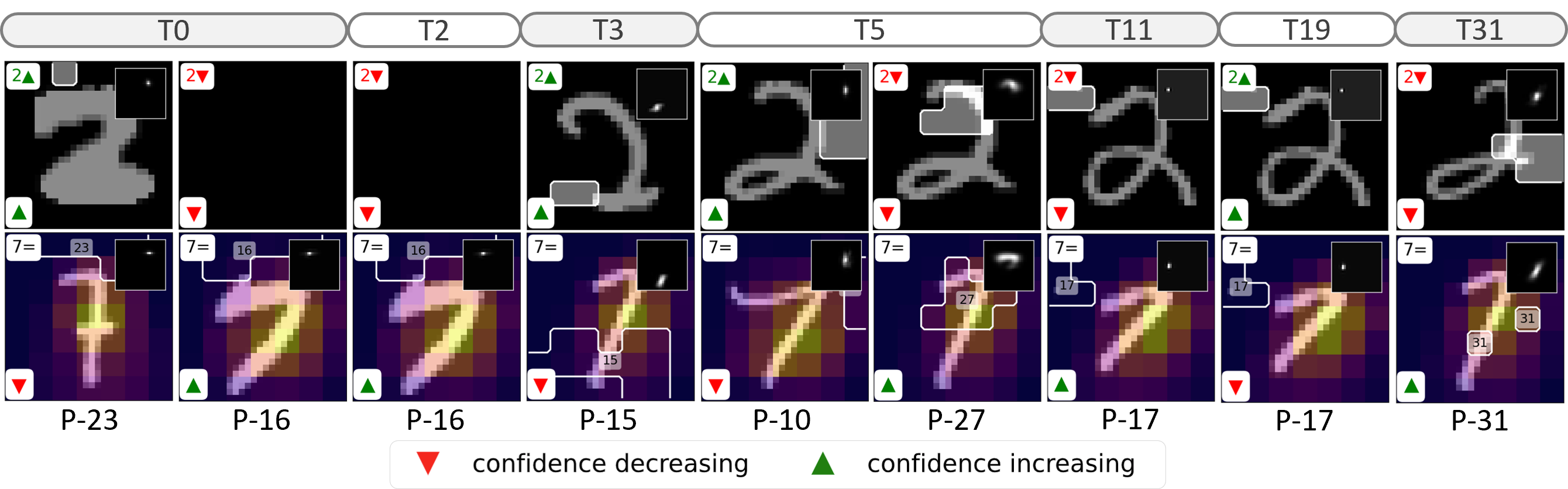}
    \caption{Visualization of prototype contributions distinguishing digit 7 from digit 2 for the secound round re-trained model. Each column shows how sharing a given prototype affects the prediction confidence for digit 7 and indicates which other digit is responsible for the confidence change at each training checkpoint.}
    \label{fig:tlcmp:2_7:t1}
\end{figure}

To investigate how the variation in training data order influences model predictions and inference strategies, we re-trained the model from scratch and compared the structured explanations for distinguishing digits 2 and 7 across two training rounds.
As shown in~\cref{fig:split:2_7:t1} and~\cref{fig:tlcmp:2_7:t1}, the two training runs adopt different prototype-based strategies. 
In the first run, the model begins with prototype P-20, corresponding to the center-right corner of digit 7. 
In contrast, the second run starts with P-16, a dot at the upper-center, resulting in a distinct strategy for separating digits 2 and 7. 
This shift introduces an uncertain prototype, P-17, in the second training round.
\begin{figure}[t]
    \centering
    \includegraphics[width=\linewidth]{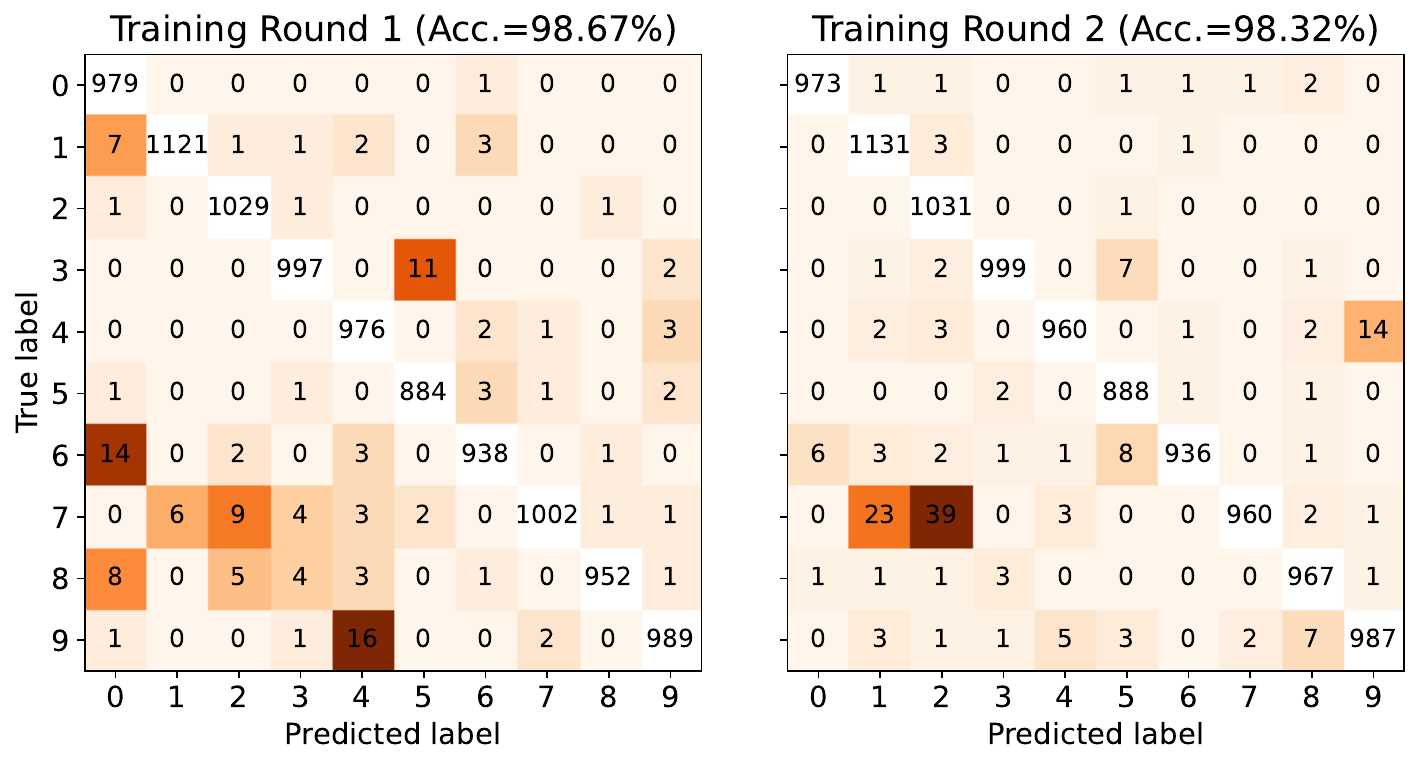}
    \caption{Difference of confusion matrices between the two training rounds evaluated on the test dataset. 
    Due to the distinct training strategies, the second training introduces an uncertain prototype that increases confusion between digits 7 and 2.}
    \label{fig:cm:raw}
\end{figure}
During the first run, although P-26 alternates between digits 2 and 7, each fluctuation is supported by another prototype (P-19) that reinforces confidence in digit 7. 
In the second round, however, P-17 fluctuates without supporting prototypes, causing the model to be confused by this prototype, which is shared between digits 2 and 7. 
This behavior is also reflected in the confusion matrices shown in~\cref{fig:cm:raw}. 
Evaluating both models on the test set, the second model exhibits more misclassifications from 7 to 2, highlighting the impact of training data order on learned inference strategies.

\subsection{Using Structured Explanations to Guide Fine-tuning}
Based on the previous analysis, two key questions remain: 
(1) Does the identified uncertain prototype truly affect the evolution of the model? 
(2) Can we guide the model to mitigate the influence of this uncertain prototype to achieve better performance and more effective optimization?

To validate this finding and demonstrate how structured explanations can guide model training, we design the following experiments: 
(1) fine-tuning the model on the entire explanation dataset (10\% of the training data), 
(2) fine-tuning on a curated dataset selected from the explanation dataset by excluding samples containing the identified uncertain prototype, and 
(3) sequential fine-tuning on the curated dataset followed by the full dataset, simulating a one-epoch perturbation during standard training. 
Each experiment is repeated 5 times to compute the average performance to reduce the randomness affection.
As identified in~\cref{fig:tlcmp:2_7:t1} and~\cref{fig:split:2_7:t1}, the uncertain prototype is P-17. Therefore, we construct the curated dataset with samples without prototype P-17 from the explanation dataset.

\begin{table}[ht]
\centering
\caption{Accuracy (\%) and confusion statistics between digits 2 and 7 after fine-tuning on the full explanation dataset (10\% of the training data) and the curated dataset derived from structured explanations. Here, $2\to7$ denotes samples with true label 2 predicted as 7, $7\to2$ denotes the opposite, and $2\leftrightarrow7$ represents their sum.}
\label{tab:cm}
\begin{tabular}{lllll}
\toprule
Dataset & Acc. (\%) & $7\to2$ & $2\to7$ & $2\leftrightarrow7$ \\
\midrule
full 
& 98.52 $\pm$ 0.34 
& 11.40 $\pm$ 8.69 
& 3.40 $\pm$ 4.03 
& 14.80 $\pm$ 6.31 \\

curated 
& \textbf{98.67 $\pm$ 0.18 }
& \textbf{5.60 $\pm$ 3.26 }
& 4.20 $\pm$ 3.19 
& \textbf{9.80 $\pm$ 2.93} \\

\midrule
full $\to$ full
& 98.46 $\pm$ 0.31 
& 7.60 $\pm$ 3.38 
& 2.00 $\pm$ 2.53 
& 9.60 $\pm$ 2.87 \\

curated $\to$ curated
& 98.31 $\pm$ 0.63 
& \textbf{6.40 $\pm$ 5.95} 
& 2.60 $\pm$ 2.42 
& \textbf{9.00 $\pm$ 4.90} \\

curated $\to$ full
& \textbf{98.64 $\pm$ 0.12} 
& \textbf{4.80 $\pm$ 2.79} 
& 3.60 $\pm$ 3.44 
& \textbf{8.40 $\pm$ 1.85} \\

\bottomrule
\end{tabular}
\end{table}

As shown in~\cref{tab:cm}, fine-tuning for one epoch on the curated dataset significantly reduces confusion between digits 2 and 7 compared to fine-tuning on the full dataset (from 14.80 to 9.80). 
However, after two epochs of fine-tuning, although the curated dataset still yields lower confusion between digits 2 and 7, the standard deviation increases from 2.93 to 4.90, which is nearly twice that of fine-tuning on the full dataset (2.87). 
This instability may arise because the model begins to explore new prototypes or alternative inference strategies not covered in the prior explanation analysis, and the biased curated dataset negatively affects this adaptation process.

To address this issue, we further fine-tune the model for one epoch on the curated dataset followed by one epoch on the full dataset. 
The results indicate that this perturbation strategy enables the model to bypass the adverse influence of P-17 while continuing optimization toward the desired objective. 
It achieves lower but more stable confusion between digits 2 and 7 ($8.40 \pm 1.85$) along with higher overall accuracy (98.64\%).

\subsection{Generalization to CIFAR10 and InceptionV3}
To demonstrate that I2X is a generalizable framework for constructing structured explanations across different domains and architectures, we further evaluate it on ResNet-50 trained on CIFAR10 and InceptionV3 trained on MNIST. 
For the CIFAR10 experiment, the number of clusters in feature K-Means clustering is set to 128 to capture the greater diversity of visual patterns.

\begin{figure}[t]
    \centering
    \includegraphics[width=0.48\linewidth]{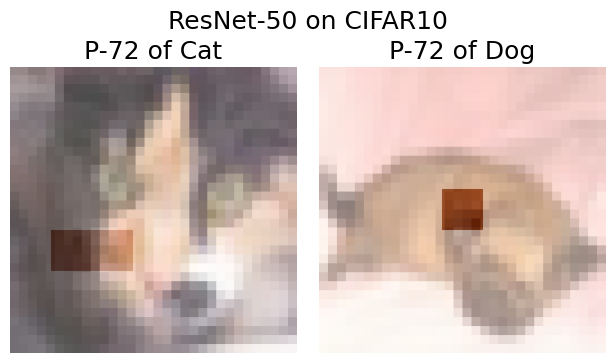}
    \includegraphics[width=0.48\linewidth]{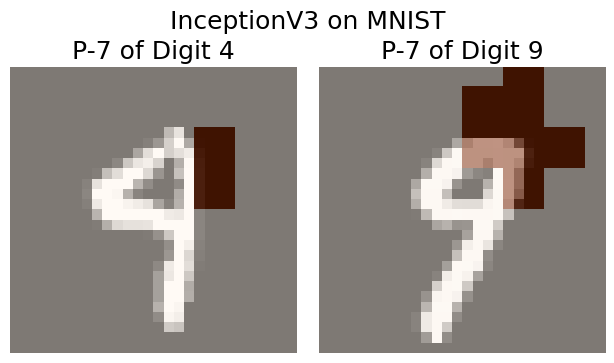}
    \caption{Identified uncertain prototypes in ResNet-50 on CIFAR10 and InceptionV3 on MNIST. Prototype P-72 (ResNet-50, CIFAR10) captures the edge between black and orange regions, leading to confusion between orange cats and dogs. Prototype P-7 (InceptionV3, MNIST) corresponds to the upper-right arc of digit 9, which resembles certain handwritten forms of digit 4, thereby causing misclassification.}
    \label{fig:proto:general}
\end{figure}

As shown in~\cref{fig:proto:general}, the most influential uncertain prototype in CIFAR10 is P-72, which corresponds to the edge between black and orange regions. 
This prototype contributes significantly to confusion between cats and dogs, particularly for orange-colored cats and dogs with similar edge structures. 
As shown in~\cref{tab:cm:general}, fine-tuning the model after removing cat and dog samples containing P-72 reduces confusion (238.60) and improves overall accuracy (84.02).

\begin{table}[t]
\centering
\setlength{\tabcolsep}{4.5pt}
\caption{Accuracy (\%) and confusion statistics between objects cat and dog or digit 4 and 9 after fine-tuning on the full explanation dataset (10\% of the training data) and the curated dataset derived from structured explanations.}
\label{tab:cm:general}
\begin{tabular}{l|ll|ll}
    \toprule
    &
    \multicolumn{2}{|l}{\textit{ResNet-50 on CIFAR10}} &
    \multicolumn{2}{|l}{\textit{InceptionV3 on MNIST}} \\
    \midrule
    Dataset & Acc. (\%)  & cat$\leftrightarrow$dog & Acc. (\%) & $4\leftrightarrow9$\\
    \midrule
    full $\to$ full
    & 81.43 $\pm$ 2.79
    & 261.20 $\pm$ 30.77
    & 99.13 $\pm$ 0.29
    & 12.60 $\pm$ 3.07 \\
    
    curated  $\to$ full
    & \textbf{84.02 $\pm$ 2.70} 
    & \textbf{238.60 $\pm$ 21.90}
    & 99.11 $\pm$ 0.27
    & \textbf{10.80 $\pm$ 2.71}\\

    \bottomrule
\end{tabular}
\end{table}

A similar phenomenon is observed for InceptionV3 on MNIST. 
Digits 7 and 4 are confused due to prototype P-7, which represents the upper-right arc of digit 9. 
Some handwritten digit 4 samples resemble digit 9 with two strokes closing at the top, activating P-7 and leading to misclassification. 
Fine-tuning without these samples reduces the confusion count from 12.60 to 10.80.
These results validate the effectiveness of I2X in extracting structured explanations across models and datasets. 

    
    


    \section{Conclusion}
In this paper, we propose the Interpretation to Explanation (I2X) framework to analyze how deep neural networks organize and evolve prototype evidence during training. By mapping the relationship between confidence and prototype intensity changes across checkpoints, I2X transforms unstructured interpretations into structured explanations that can reveal which prototypes contribute to predictions and how they evolve during optimization.


With I2X analysis, we can understand how the model distinguishes a digit from others, typically starting with digits supported by certain prototype and then resolving more uncertain cases.
Additionally we show with experiments on MNIST and CIFAR10 with ResNet50 and InceptionV3 that I2X that variation in training data order can be observed in the prototype selection sequences and inference strategies, with quantitative changes to inter- and intra-class confusion. Finally, we show how I2X can be used to design a fine-tuning strategy in which selectively perturbing uncertain prototypes reduces confusion and improves performance and stability. Overall, I2X provides a practical framework for understanding model learning and guiding data-centric model optimization.
                                       
In future work we hope to integrate I2X with explain-by-design approaches such as prototype learning (\eg, ProtoPNet~\cite{chen2019looks}) to reduce the cost of its current reliance on post-hoc methods and clustering. We also hope to generalize the concept of uncertainty in prototypes (e.g. with a metric as in~\cite{finzi2026entropy}) to potentially measure and guide improved performance of training and model construction. 
    \section*{Acknowledgements}
The authors acknowledge support from the Harold L. and Heather E. Jurist Center of Excellence for Artificial Intelligence at Tulane University.
    
    %
    %
    \bibliographystyle{styles/splncs04}
    \bibliography{main}
\end{document}